%% file: main.tex
\documentclass{article}

\PassOptionsToPackage{numbers, compress}{natbib}

    \usepackage[preprint]{neurips_2024}

\usepackage{listings}
\usepackage[utf8]{inputenc} % allow utf-8 input
\usepackage[T1]{fontenc}    % use 8-bit T1 fonts
\usepackage[colorlinks, linkcolor=orange, anchorcolor=blue, citecolor=blue]{hyperref}
\usepackage{url}            % simple URL typesetting
\usepackage{booktabs}       % professional-quality tables
\usepackage{amsfonts}       % blackboard math symbols
\usepackage{nicefrac}       % compact symbols for 1/2, etc.
\usepackage{microtype}      % microtypography
\usepackage{xcolor}         % colors
\usepackage{graphicx,float,wrapfig}
\usepackage{subfigure}
\usepackage{wrapfig}
\usepackage{multirow, multicol}
\usepackage{caption}
\usepackage{fontawesome5}
\usepackage{url}
\input{math_commands}

\input{symbols}

\usepackage{enumitem}
\setlist[itemize]{leftmargin=*}
\setlist[enumerate]{leftmargin=*}

\def\RD{{\tt{RNAdiffusion}} }

\newcommand{\enc}{\mathrm{Enc}}
\newcommand{\qf}{\mathrm{QFormer}}
\newcommand{\dec}{\mathrm{Dec}}
\newcommand{\xb}{\boldsymbol{x}}
\renewcommand{\zb}{\boldsymbol{z}}
\newcommand{\dm}{\mathrm{Denoiser}}
\newcommand{\EE}{\operatorname{\mathbb{E}}}

\newcommand\codeurl[1]{{{\color{blue}{\url{#1}}}}}

\title{Latent Diffusion Models for Controllable RNA Sequence Generation}

\author{
Kaixuan Huang$^{1}$\thanks{Equal contribution. $^{\natural}$ Intern at Stanford University.}
\;
Yukang Yang$^{1*}$
\;
Kaidi Fu$^{2\natural}$
\;
\textbf{Yanyi Chu}$^{3}$
\;
\textbf{Le Cong}$^{3}$
\;
\textbf{Mengdi Wang}$^{1}$\thanks{Corresponding author: mengdiw@princeton.edu}
\\
\vspace{4mm}
$^{1}$Princeton University \;
$^{2}$Tsinghua University \;
$^{3}$Stanford University
\\
}

\begin{document}
\maketitle
\begin{abstract}
  \input{sections/abstract}
\end{abstract}

%\clearpage

\input{sections/intro}

\input{sections/method}

\input{sections/experiment}

\input{sections/related_work}

\input{sections/conclusion}

\clearpage

\input{sections/acknowledgement}

\bibliographystyle{plainnat}
\bibliography{ref}

\newpage
\appendix

\clearpage
\input{appendix/limitation}

\input{appendix/impact}
\input{appendix/related_work}
\input{appendix/exp}

%\input{appendix/nll.tex}

%\input{appendix/exp_log/exp_log_rna}
% \input{appendix/exp_log/exp_log_text}
% \input{appendix/exp_log/exp_log}

% \clearpage
% \input{sections/checklist.tex}

\end{document}

%% file: math_commands.tex
\usepackage{amsmath,amsfonts,bm}

\def\eqref#1{equation~\ref{#1}}
\def\1{\bm{1}}

\DeclareMathAlphabet{\mathsfit}{\encodingdefault}{\sfdefault}{m}{sl}
\SetMathAlphabet{\mathsfit}{bold}{\encodingdefault}{\sfdefault}{bx}{n}

%% file: symbols.tex
\usepackage{enumitem}
\usepackage{comment} 
\usepackage{hyperref}
\usepackage{natbib}
\usepackage{bm}

\usepackage{thmtools}
\usepackage{thm-restate}
\usepackage{times}

\usepackage{amssymb}
\usepackage{amsmath}
\usepackage{amsfonts}
\usepackage{amsthm}

\usepackage[utf8]{inputenc} % allow utf-8 input
\usepackage[T1]{fontenc}    % use 8-bit T1 fonts
\usepackage{url}            % simple URL typesetting
\usepackage{booktabs}       % professional-quality tables
\usepackage{amsfonts}       % blackboard math symbols
\usepackage{nicefrac}       % compact symbols for 1/2, etc.
\usepackage{microtype}      % microtypography
\usepackage{xcolor}

\def\approxcorrect{\cmark\kern-1.4ex\raisebox{.30ex}{$\xmark$}}

\newcommand{\idxn}[1][]{\ifthenelse{\equal{#1}{}}{\mathbb{INDQ}_n}{\mathbb{INDQ}_{#1}}}

\newcommand{\beq}{\begin{equation}}

\newcommand{\eeq}{\end{equation}}
\newcommand{\zb}{{\bar{\z}}}

\def \endprf{\hfill {\vrule height6pt width6pt depth0pt}\medskip}

%% file: sections/abstract.tex
This work presents \RD, a latent diffusion model for generating and optimizing discrete RNA sequences of variable lengths.
RNA is a key intermediary between DNA and protein, exhibiting high sequence diversity and complex three-dimensional structures to support a wide range of functions. 
We utilize pretrained BERT-type models to encode raw RNA sequences into token-level, biologically meaningful representations. A Query Transformer is employed to compress such representations into a set of fixed-length latent vectors, with an autoregressive decoder trained to reconstruct RNA sequences from these latent variables.
We then develop a continuous diffusion model within this latent space. 
To enable optimization, we integrate the gradients of reward models—surrogates for RNA functional properties—into the backward diffusion process, thereby generating RNAs with high reward scores.
Empirical results confirm that \RD generates non-coding RNAs that align with natural distributions across various biological metrics. 
Further, we fine-tune the diffusion model on mRNA 5’ untranslated regions (5’-UTRs) and optimize sequences for high translation efficiencies.
Our guided diffusion model effectively generates diverse 5'-UTRs with high Mean Ribosome Loading (MRL) and Translation Efficiency (TE), outperforming baselines in balancing rewards and structural stability \textit{trade-off}.
Our findings hold potential for advancing RNA sequence-function research and therapeutic RNA design.

%% file: sections/intro.tex
\section{Introduction}
\label{sec:intro}

Diffusion models demonstrate exceptional performances in modelling continuous data, with applications in images synthesis ~\citep{ramesh2022hierarchical, rombach2022high, balaji2022ediffi}, point clouds generation ~\citep{nichol2022point}, video synthesis ~\citep{ho2022imagen}, reinforcement learning ~\citep{ajay2023is, janner2022diffuser, liang2023adaptdiffuser}, time series ~\citep{tashiro2021csdi} and molecule structure generation ~\citep{Watson2022.12.09.519842}. 
An important advantage of diffusion models is that their generation process can be ``controlled" to achieve specific objectives via incorporating additional \textit{guidance} signal. The guidance can steer the backward process toward generating samples with desired properties, without additional training~\citep{dhariwal2021diffusion, bansal2023universal, chung2022diffusion}.

While diffusion models have been widely applied in continuous domains, they have also proven useful in generating discrete sequences, inheriting the benefits of controllability through guidance. This has been shown both in the textual domain~\citep{austin2021structured, li2022diffusion} and in the biology domain~\citep{ferreira2024dna, Watson2022.12.09.519842, Luo2022.07.10.499510, gruver2023protein}.
In the domain of text generation, state-of-the-art performances are achieved by scaling up auto-regressive models~\citep{achiam2023gpt, anil2023palm, team2023gemini, touvron2023llama, touvron2023llama2}. However, for generating biological sequences such as proteins, DNA, and RNA, language models have to be adapted to handle their specific sequence characteristics~\citep{rao2021msa,jumper2021highly, nijkamp2023progen2, benegas2023dna, ji2021dnabert, brandes2022proteinbert}.

In this work, we focus on developing generative models for RNA, with applications to both non-coding RNAs (ncRNAs) and mRNA untranslated regions (UTRs).

For ncRNAs, they are crucial regulators of gene expression and cellular processes, influencing pathways involved in development, disease, and immune responses. In parallel, the UTRs of mRNAs play a central role in regulating gene expression~\citep{miao2017rna,caprara2000rna,atkins2011rna} and mRNA stability~\citep{araujo2012before}. 
The ability to generate optimized UTR sequences offers significant potential for controlling gene expression, advancing mRNA vaccine designs, and enabling targeted therapies~\citep{sample2019human,cao2021high,barazandeh2023utrgan}. 
Therefore, our goal is to develop a robust and versatile language model capable of generating ncRNAs, and optimizing UTRs for enhanced functionality.

\begin{figure}[t]
    \centering
    \includegraphics[width=0.95\textwidth]{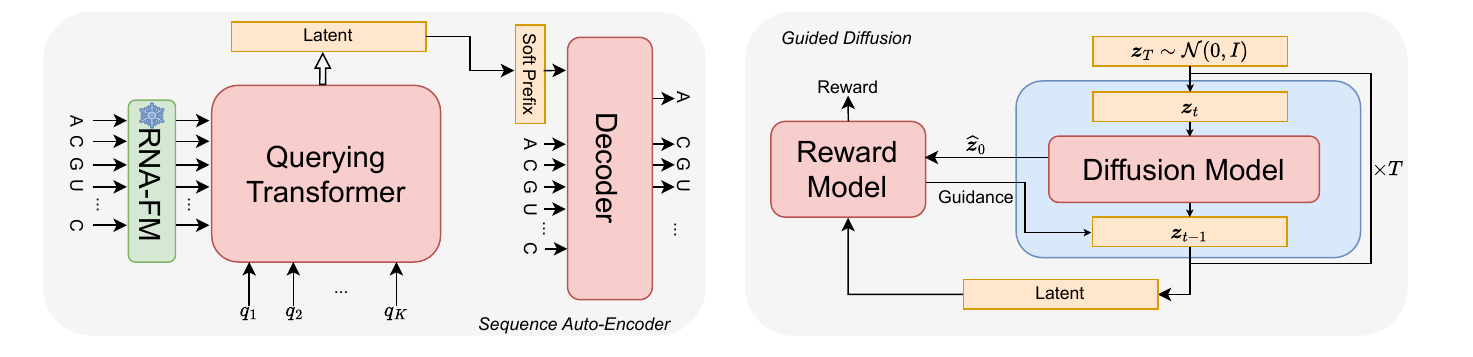}
    \caption{\RD: Latent diffusion model for RNA sequences. Three parts of \RD: (1) RNA \textbf{sequence auto-encoder}, consisting of a pretrained RNA-FM model, a Querying Transformer, and a decoder, for translating between the sequence space and the latent space; (2) \textbf{Guided diffusion} model with a pre-trained score network, for generating latent RNA embeddings under external guidance; (3) \textbf{Latent reward model}, trained on the latent space to predict functional properties of RNA, for computing guidance of diffusion.  
    } 
    \vspace{-2em}
    \label{fig:main}
\end{figure}

For ncRNAs and UTRs, individual nucleotides (A, U, G, C) often do not independently exhibit functions. Instead, the nucleotides may form clusters of higher-order structures, such as motifs, contributing to the functional properties of the RNA. This introduces distinct challenges in sequence modeling.
Specifically, the denoising network has to account for the high-level picture of the entire sequence and the low-level syntax within a local cluster. Besides, these RNAs have dramatically different lengths, adding difficulty to efficient training and generalization. 
Although Transformer denoisers can handle variable-length inputs in theory, existing studies mostly pad or truncate input to a fixed length, or build an additional length prediction module to ensure robust practical performances~\citep{savinov2021step, johnson2021beyond, ye2023diffusion}. 

To overcome these challenges, we develop diffusion models on a latent space for RNA sequences. Inspired by recent advances~\citep{rombach2022high, lovelace2022latent}, our approach utilizes a sequence auto-encoder to map raw input sequences into a latent embedding space. We then train a score-based continuous diffusion model to learn this latent distribution. Please see Figure~\ref{fig:main} for a detailed schematic. This latent-space approach is effective for RNA sequences: \textit{First}, it delegates the generation of the intricate raw RNA sequences to the sequence decoder, allowing the diffusion model to concentrate on capturing the coarse-grained distribution in the meaningful latent space. \textit{Second}, the latent space is of fixed size, simplifying the design and training of the diffusion model by eliminating the need for additional modules or special adaptations for varying sequence lengths.

For the sequence autoencoder, we use a pretrained language model RNA-FM~\citep{chen2022interpretable} to map raw sequences to meaningful, token-level representations. A trainable Query Transformer (Q-Former)~\citep{li2023blip} summarizes these representations into fixed-length latent vectors.
Next, we train a score-based denoising diffusion model to capture the distribution of RNAs in latent space. We then assess the generated ncRNAs using biological metrics such as Levenshtein distance, 4-mer distance, G/C content, minimum free energy, and secondary structure. Results confirm that the generated ncRNAs align with natural ncRNA distributions across these indicators. Thus, our latent diffusion model can produce novel RNA sequences with high similarity to native ncRNAs.

Since latent variables from Q-Formers provide biologically meaningful summaries, we train reward networks directly in the latent space to predict functional properties of UTRs, key elements in mRNAs. The rewards, Mean Ribosome Loading (MRL)~\citep{sample2019human} and Translation Efficiency (TE)~\citep{cao2021high}, measure mRNA-to-protein production levels and efficiencies, respectively. Using the trained reward models, we compute gradients and guide the diffusion model. Results show that {\tt{RNAdiffusion}} generates novel UTRs with substantially higher MRL and TE values by incorporating guidance into the backward generation process. This guidance is a plug-and-play method that requires no extra training.
Overall, the generated UTRs from guided diffusion achieve a 166.7\% improvement of TE and a 52.6\%  improvement of MRL compared to unguided generation.

We summarize the contributions of \RD:

\begin{itemize}[itemsep=1pt, parsep=1pt, topsep=2pt]
    \item 
     Our sequence autoencoder, a novel adaptation of Q-Former, summarizes varying-length outputs from pretrained encoders into fixed-length embeddings, accurately reconstructing input sequences.
    \item Our latent diffusion model generates novel RNA sequences, and empirical studies confirm that these sequences align with natural distributions across various biological metrics.
    \item \RD enables \textbf{controlled generation} via guidance, optimizing 5'-UTRs for higher protein production ($166.7\%$ improved TE and $52.6\%$ improved MRL compared to unguided generation), while achieving a better reward–structural stability \textit{trade-off} than baselines.
    
\end{itemize}

%% file: sections/method.tex
\section{Methodology}

\RD has three parts: (1) a sequence auto-encoder and decoder that map between the space of variable-length discrete sequence and a continuous latent space, (2) a continuous diffusion model trained to model the distributions in the latent space, with optionally added guidance, and (3) a reward model trained on top of the latent space for computing guidance. See Figure~\ref{fig:main} for an overview of our methodology.

\subsection{Sequence Auto-Encoder}

We aim for a sequence auto-encoder that removes the information redundancy in the raw sequence of tokens and compress them into a fixed-length set of continuous embedding vectors. 
We propose to use a sequence auto-encoder that consists of a pretrained encoder, a Querying Transformer (Q-Former)~\citep{li2023blip}, and a decoder (Figure~\ref{fig:main} left panel). 

\textbf{Encoder.} We adopt the pretrained RNA-FM~\citep{chen2022interpretable}, a BERT-type encoder-only model, as the first component of the sequence auto-encoder~\citep{devlin-etal-2019-bert}. BERT-type encoders are pretrained via masked language modeling task to predict randomly masked tokens. We denote the encoder by $\enc$. It maps a sequence $\xb = (\xb^1, \xb^2, \dots, \xb^L)$ to a sequence of contextualized token embeddings of the same length $L$.  We fix the encoder during the training of the sequence auto-encoder.

\textbf{Querying-Transformer.} Next we use a Q-Former, denoted by $\qf_{\phi}(\cdot)$, to summarize the embeddings of the pre-trained encoder $\mathrm{Enc}(x)$ into $K$ latent vectors $\zb = (\zb^1,\zb^2,\dots, \zb^K)$, where $K$ is a 
\textit{fixed} number independent of $L$. 
We aim to achieve three goals:  (1) To retains enough information in latent vectors, making it easy to train a decoder for raw sequence reconstruction; (2) To use latent vectors later for predicting RNA functional properties, via training additional prediction models on top of these embedding;
 (3) To reduce the dimensionality of raw sequences, so that diffusion models can focus on learning intrinsic structure of latent RNA vectors. 

The Querying Transformer (Q-Former)~\citep{li2023blip} takes $K$ trainable query token embeddings $(\boldsymbol{q}^1, \dots, \boldsymbol{q}^K)$ as input. The query tokens go through the transformer with cross-attentions to the embedding sequence given by the encoder and progressively summarize the original token embeddings of varying lengths into a fixed-size sequence of embedding vectors. 
This is a flexible approach compared to using only the \verb|<cls>| token embedding ~\citep{li-etal-2020-optimus}. 
As when we increase $K$, we retain more information in the latent vectors and allow accurate reconstruction, which helps to mitigate the KL-vanishing problem~\citep{bowman-etal-2016-generating}. Our ablation study in Section~\ref{sec:autoencoder:abalation} validates the necessity of the Q-former.

\textbf{Decoder.}
We use a standard causal Transformer~\citep{vaswani2017attention} as the decoder, denoted by $\dec_\psi$. We build a simple linear projection layer to convert the $K$ latent vectors into a soft prompt embeddings of equal size for the decoder to condition on. Then the decoder autoregressively reconstructs the original input sequence given the latent vectors. 

\textbf{Traning Objective.}
With the pre-trained encoder fixed, we train the Q-Former and the decoder from scratch in an end-to-end manner. 
The training objective is to reconstruct the original sequence from the embeddings given by the fixed pretrained encoder model.  Our loss function is given by
% \[
\begin{equation}\label{eq:ae_loss}
    \mathcal{L}_{\text{reconstruct}}(\psi, \phi ;\xb) =  -\log p_{{\dec}_\psi} (\xb^1,\xb^2,\dots,\xb^L \mid \qf_\phi(\enc(\xb))), 
% \]
\end{equation}
where $\phi,\psi$ denote the weights of the QFormer (including the query tokens $\boldsymbol{q}$) and Decoder, respectively.
As people often use a very small (e.g. $1e-6$) KL-regularization term in latent diffusion models~\citep{rombach2022high, zhang2023planner}, we choose to not include any KL-regularization, simplifying the design and helping to avoid KL-vanishing ~\citep{bowman-etal-2016-generating}. Empirically, the training works well without the KL term. 

\subsection{Latent Diffusion Model} 
\label{sec:method:dm}

\newcommand{\epsilonb}{\boldsymbol{\epsilon}}
We build a classical continuous diffusion model~\citep{ho2020denoising} to model the distribution of the latent variables $\zb$, which generates samples from the target distribution by a series of noise removal processes. Specifically, we adopt a denoising network to predict the added noise $\epsilonb$ from the noisy input $(\sqrt{\bar{\alpha}_t} \zb_0 + \sqrt{1- \bar{\alpha}_t} \epsilonb)$,  where $t$ is a time step of the forward process, $\bar{\alpha}_t$ is the corresponding signal strength level, and $\epsilonb$ is the added Gaussian noise.

For the denoising score network $\dm_\theta$, we use transformer denoising networks following~\citep{li2022diffusion}. We use the $\epsilon$-prediction scheme in Denoising Diffusion Probabilistic Models (DDPM)~\citep{ho2020denoising} and adopt the simplified training objective given by
% \[
\begin{equation}\label{eq:diff_loss}    
    \mathcal{L}(\theta) = \mathbb{E}_{\zb_0 = \qf_\phi(\enc(\xb))} \mathbb{E}_{t, \epsilonb} \|\epsilonb - \dm_\theta(\sqrt{\bar{\alpha}_t} \zb_0 + \sqrt{1- \bar{\alpha}_t} \epsilonb, t )  \|^2,
% \]
\end{equation}

In our framework, the latent embedding have fixed sizes, making it easy to train diffusion models and treating data as if they are continuous like images. Unlike other discrete diffusion models, our framework does not require truncating/padding sequences to a fixed length compared with~\citep{li2022diffusion, austin2021structured}, and it does not require training an additional length predictor as done by ~\citep{savinov2021step, johnson2021beyond, ye2023diffusion}. Our auto-regressive decoder has been trained to generate sequences with variable lengths; see Figure~\ref{fig:seq_len_nc}.

\subsection{Guided Generation in the Latent Space}
\label{sec:method:dm:guidance}

To generate RNA sequences of desired properties, we need a reward model for predicting the property of new sequences.
Suppose we have a labeled dataset $\{(\xb, r)\}$, where $r$ is the measured reward of interest, e.g., the translation efficiency. We train a reward model $R_\xi$ on top of the latent space to predict $r$ given the input sequence $\xb$. The training objective is
% \[
\begin{equation}\label{eq:reward_loss}
    \mathcal{L}(\xi) = \EE_{(\xb, r)} \ell( R_\xi(\zb) , r),
% \]  
\end{equation}
where $\ell$ is a loss function, $\zb = \qf_\phi(\enc(\xb))$ is the embedding obtained from Q-Former, and $\xi$ are weights of the reward model.

Finally, we use the reward model to compute a gradient-based guidance (e.g., via universal guidance~\citep{bansal2023universal}) and add it to the backward generation process. This guidance steers the random paths of the backward generation process of the diffusion model toward a higher reward region in a plug-and-play manner. Specifically, at the step $t$ of the backward diffusion process, given the current noisy sample $\zb_t$, we first estimate the clean sample 
% \[ 
\begin{equation}\label{eq:z0_estim}
    \widehat{\zb}_0 = \widehat{\zb}_0(\zb_t) = \frac{\zb_t - \sqrt{1-\bar\alpha_t} \dm_\theta(\zb_t, t)}{\sqrt{\bar\alpha_t}}.
% \]
\end{equation}
And we use $R_\xi (\widehat{\zb}_0 )$ as a surrogate reward for the noisy sample $\zb_t$. We plug in the gradient $\nabla_{\zb_t} \ell(r^*, R_\xi (\widehat{\zb}_0 ))$ into the current predicted noise as follows, with $r^*$ being the targeted reward 
%\mw{bad notation} 
and $\lambda$ being the guidance strength.
% \[
\begin{equation}\label{eq:pred_guidance}
    \text{predicted noise} = \dm_\theta(\zb_t, t) + \sqrt{1-\bar\alpha_t} \cdot \lambda \cdot \nabla_{\zb_t} \ell(r^*, R_\xi (\widehat{\zb}_0 )). 
% \]
\end{equation}
We note that controlling the generation for decoder-only models such as GPTs in the raw sequence space is generally hard without fine-tuning the model (see ~\citep{korbak2023pretraining} and the references therein). 

%% file: sections/experiment.tex
\section{RNA Sequence AutoEncoder}
\label{sec:exp:autoencoder}

\subsection{Experiment Setup}

\textbf{Models.} We adopt the pretrained RNA-FM~\citep{chen2022interpretable} as the encoder. RNA-FM is an encoder-only model trained with masked language modelling on non-coding RNAs (ncRNAs). The architecture of our Q-Former is modified from the ESM-2 model structure~\citep{lin2022language}. For the decoder model, we adopt the same model architecture of ProGen2-small~\citep{nijkamp2023progen2} with our customized tokenizers for RNAs. 

\textbf{Dataset.} 
We pretrain our RNA sequence autoencoder with the ncRNA subset of Ensembl database~\citep{10.1093/nar/gkac958} downloaded from RNAcentral~\citep{rnacentral2021rnacentral}, which contains collected ncRNA sequences from a variety of vertebrate genomes and model organisms. We select the sequences shorter than 768 bp to avoid exceeding RNA-FM's maximum length limit
and exclude those with rare tokens more than \{A, C, G, U\}, curating a dataset of 1.1M samples for training and testing.

\textbf{Hyperparameter Search.} There are two hyperparameters that influence the dimension of the latent space: the number of query tokens 
$K$ and the hidden dimension $D$ of each latent vector. We tune $K$ among $\{16, 32\}$ and $D$ among $\{40, 80, 160, 320\}$.

\textbf{Evaluation Metrics.} For each input sequence $\xb$ of length $L$, we calculate two metrics: (1)length-normalized reconstruction loss (i.e., negative log-likelihood, \textbf{NLL}), i.e., $-\frac{1}{L}\log p_{\dec_\psi}( \xb \mid \qf_\phi(\enc(\xb)));$
(2)length-normalized edit distance (\textbf{NED}) between the reconstructed sequence 
$\widehat{\xb}\sim p_{\dec_\psi}(\cdot\mid\qf_\phi(\enc(\xb)))$ 
and the original input sequence $\xb$, i.e., $\frac{1}{L}\mathrm{EditDistance}(\widehat{\xb}, \xb).$
Due to the space limit, additional experimental setup can be found in Appendix~\ref{appendix:exp:ae}.

\subsection{Results}
We perform a sweep over hyperparameters and obtain a set of RNA sequence autoencoders with high reconstruction capabilities. For each of the two metrics, we report the empirical average on a held-out test set in Table~\ref{tab:AE}.   
We see that increasing either the number of query tokens $K$ or the hidden dimension $D$ reduces the reconstruction errors and the error rates. Increasing the dimension of the latent space retains more information in the latent embeddings, and hence it is easier to reconstruct the input sequence.

\begin{table*}[ht]
    \centering
    \caption{The performance of the RNA sequence autoencoders and the latent diffusion models. For diffusion models, the upper bound of NLL, i.e., ELBO, is calculated using Equation (12) of \citep{ho2020denoising}. 
 }
    \resizebox{0.55\linewidth}{!}
    {
      \begin{tabular}{cc||cc|c}
      \toprule 
       \multirow{2}{*}{$K$} & \multirow{2}{*}{$D$} & \multicolumn{2}{c|}{Sequence AutoEncoder}  & Latent Diffusion \\
        &  & NLL$\downarrow$ &  NED $\downarrow$ & NLL  (bits/dim)$\downarrow$\\
        \midrule
      16 & 40 & 0.2631 & 19.36\% $\pm$ 21.76 \% & $\leq$ 2.93 \\  
      16 & 80 & 0.0186 &  6.00\% $\pm$ 13.40\% & $\leq$ 1.64 \\        
      16 & 160 & 0.0156 &  2.19\% $\pm$ 5.50 \% & $\leq$ 1.78 \\  
      16 & 320 & 0.0023 &  0.12\% $\pm$ 0.66 \% & $\leq$ 1.61 \\  
      \midrule
        32 & 40 & 0.0995 & 13.50\% $\pm$ 19.06 \% & $\leq$ 2.35 \\  
        32 & 80 & 0.0125 & 4.40\% $\pm$ 10.73 \%  & $\leq$ 1.29 \\  
        32 & 160 & 0.0003 & 0.02\% $\pm$ 0.40 \% & $\leq$ \textbf{1.01} \\  
        32 & 320 & \textbf{0.0001} & \textbf{0.003\% $\pm$ 0.08 \%} & $\leq$  1.68 \\  
      \bottomrule 
      \end{tabular} 
      }
    
    \label{tab:AE}% 
\end{table*}%

\subsection{Ablation Studies.}
\label{sec:autoencoder:abalation}
We perform ablation studies to validate the necessities of the components of our sequence auto-encoder.
When we replace the Q-former with a naive average pooling of RNA-FM output token embeddings $\mathrm{Enc}(x)$, the reconstruction performance of the Sequence AutoEncoder is significantly dampened, resulting in a 34.8\% normalized edit distance (see Table~\ref{tab:ab_qformer}). 

We also experiment with replacing the RNA-FM with naive one-hot token embeddings and the results indicate that both the reconstruction capability of the sequence autoencoder and the reward modeling ability of the latent vectors are negatively affected, which validate the necessity of using the pre-trained RNA-FM as the encoder. Full results are deferred to Appendix~\ref{appendix:exp:ab_rnafm}.

\begin{table}[h]
    \centering
    \caption{\textbf{Ablations on Q-former}. We remove the Q-former by averaging all output token embeddings (dim: 640) of RNA-FM, i.e., latent space dim $1\times640$, and train the decoder for one epoch.
    }
    \vspace{2mm}
    \resizebox{0.5\linewidth}{!}
    {
      \begin{tabular}{r||cc}
      \toprule  
      \multirow{2}{*}{
      Settings ($K\times{D}$)
      } & \multicolumn{2}{c}{Sequence AutoEncoder} \\
    & NLL $\downarrow$ &  NED $\downarrow$ \\
        \midrule  
  w/o Q-former ($1\times640$) & 1.0061  &  34.78 \% $\pm$ 17.81 \%  \\  
  w/ Q-former  ($16\times160$)  & \textbf{0.0156} & \textbf{2.19\% $\pm$ 5.50 \%}  \\ 
      \bottomrule 
      \end{tabular} 
      }
    \label{tab:ab_qformer}% 
\end{table}%

\section{Latent Diffusion Model for ncRNAs}
\label{sec:exp:ldm}

\subsection{Experimental Setup}
\label{sec:exp:eval}
\textbf{Models and Datasets.} We build a 24-layer Transformer model~\citep{vaswani2017attention} with hidden dimension $2048$ as the denoising network for our latent diffusion model. As the latent vectors are continuous, we do not need any additional embedding layer or rounding operation. We use the same Ensembl ncRNA dataset and apply the same preprocessing as our RNA sequence Auto-encoder. We trained the diffusion model for 10 epochs with the Sequence Autoencoder frozen. Full training details are given in Appendix~\ref{appendix:exp:dm}.

\textbf{Evaluation Metrics.}
%\mw{are they reported in Table 1? or where?}. \kaixuan{Yes, see Next section!}
%
% \TODO{make sure the metric description matches the newest results: (1) random seq generation (2) length-normalization. }
% \TODO{Add k-mer frequency statistics; }
%
%
To evaluate the performance of our latent diffusion model, we calculate the weighted loss $\mathcal{L}_{1:T-1}$ on the test dataset according to Eqn. (12) of~\citep{ho2020denoising}, which is an upper bound of the negative log-likelihood (NLL). We also visualize RNAs in the latent space by using a \textbf{t-SNE map}~\citep{van2008visualizing}, and compare with reference natural RNAs 
% \mw{held out?} 
and randomly generated sequences, which match the length distribution of natural data.
Besides, we feed the generated latent vectors through the decoder to get the RNA sequences and we adopt the following biological metrics following~\citep{ozden2023rnagen, barazandeh2023utrgan}.

\begin{itemize}[itemsep=1pt, parsep=1pt, topsep=2pt]
    \item \textbf{Minimum Levenshtein Distance}. The Levenshtein distance~\citep{levenshtein1966binary} is defined as the minimum number of edits (insertions, deletions, or substitutions) required to transform one sequence into another. 
    For each generated RNA sequence, 
    we calculate the smallest Levenshtein distance 
    to all sequences in a reference dataset of natural RNAs, and normalized the distance by the length of the generated sequence.
    \footnote{We exclude the identical sequence when calculating the minimum Levenshtein distance and minimum 4-mer distance.} By looping over all the generated RNA sequences, we obtain a distribution of minimum \textit{normalized} Levenshtein distances.
    
    \item \textbf{Minimum 4-mer Distance}. We calculate the frequencies of 4-mers of each sequence~\citep{vinh2015two}. The 4-mer distance of two sequences is defined as the Euclidean distance between the two \textit{frequency} vectors. 
    Similarly, for each generated RNA sequence, we calculate the minimum 4-mer distance against a reference natural RNA dataset.\footnotemark[\value{footnote}] We obtain the distribution of minimum 4-mer distances for the generated RNA sequences. 
    
    \item \textbf{G/C content}~\citep{konu2002correlations}. G/C content, defined as the percentage of guanine (G) and cytosine (C), is an essential indicator of molecular stability. A higher G/C content generally implies higher stability due to the stronger hydrogen bonds between G and C compared to A and T. 
    
    \item \textbf{Minimum Free Energy (MFE)}~\citep{trotta2014normalization}. It represents the minimum energy required for the molecule to maintain its conformation. The MFE is crucial measure of an RNA segment's structural stability, reflecting how tightly the RNA molecule can fold based on its nucleotide composition and sequence arrangement. A lower MFE suggests a more stable and tightly folded RNA structure. We calculate the MFE of generated RNA sequences using the  ViennaRNA Package~\citep{lorenz2011viennarna}. 

    \item \textbf{Minimum Levenshtein Distance of RNA Secondary Structure}. 
    RNA secondary structure, which involves local interactions like hairpins and stem-loops, plays a crucial role in the RNA's biological functions~\citep{higgs2000rna}. Dot-bracket notation represents these structures by using dots for unpaired nucleotides and matching parentheses for base pairs, simplifying the visualization and analysis of RNA sequences~\citep{lorenz2011viennarna}. We calculate the secondary structure of RNA sequences using ViennaRNA Package~\citep{lorenz2011viennarna}, and then calculate the distribution of the minimum \textit{normalized} Levenshtein distances of the secondary structures of the generated RNAs to those of natural referential RNAs.
\end{itemize}

\subsection{Results: Samples from \RD Resemble Natural ncRNAs}
\label{sec:results_nc}
\vspace{-4pt}
We report the test NLL of the latent generated distribution in Table~\ref{tab:AE}. We see that increasing the query token length $K$ and the embedding dimension $D$ generally decreases the test NLL per dimension. 
For subsequent experiments, we choose to use the trained RNA auto-encoders that have average normalized edit distance $\leq 1\%$. They correspond to hyperparameters 
$(K,D)=(16,320)$, $(32,160)$, and $(32,320)$. 
We further select 
the parameter $\textbf{(32, 160)}$ for generation, which achieves the smallest \textit{total} NLL, i.e, NLL per dimension \textit{multiplied by} the dimension of latent space ($K\times{D}$). 

\begin{wrapfigure}[12]{r}{0.29\textwidth}
\centering
\vspace{-15pt}
\includegraphics[width=0.29\textwidth]{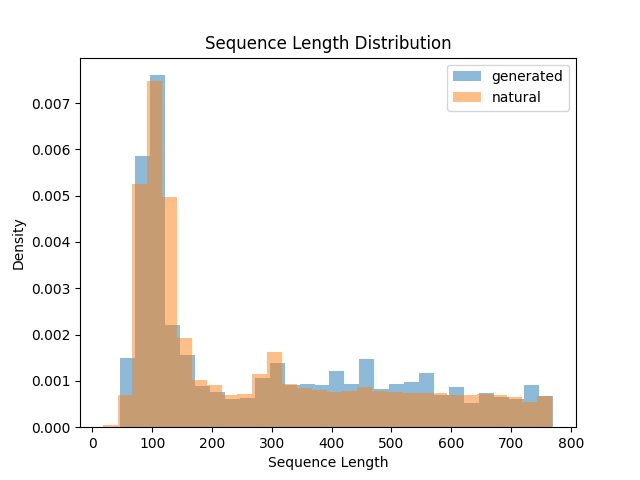}
\vspace{-12pt}
\caption{
\textbf{Sequence length comparison} between the natural \textbf{ncRNA} test set and generated sequences (sample size: 20000).
}
% }
\vspace{-60pt}
\label{fig:seq_len_nc}
\end{wrapfigure}

We compare the generated samples (with $K=32$, $D=160$) with natural ncRNAs. We plot the length distribution of the sequences in Figure~\ref{fig:seq_len_nc} and demonstrate that the generated sequences vary in length, with a distribution closely aligned with that of natural ncRNAs. 
\captionsetup[figure]{font=footnotesize,labelfont=footnotesize}
\captionsetup[figure]{font=normalsize,labelfont=normalsize}

\begin{figure}[htbp]
  \centering
  \includegraphics[width=0.85\linewidth]{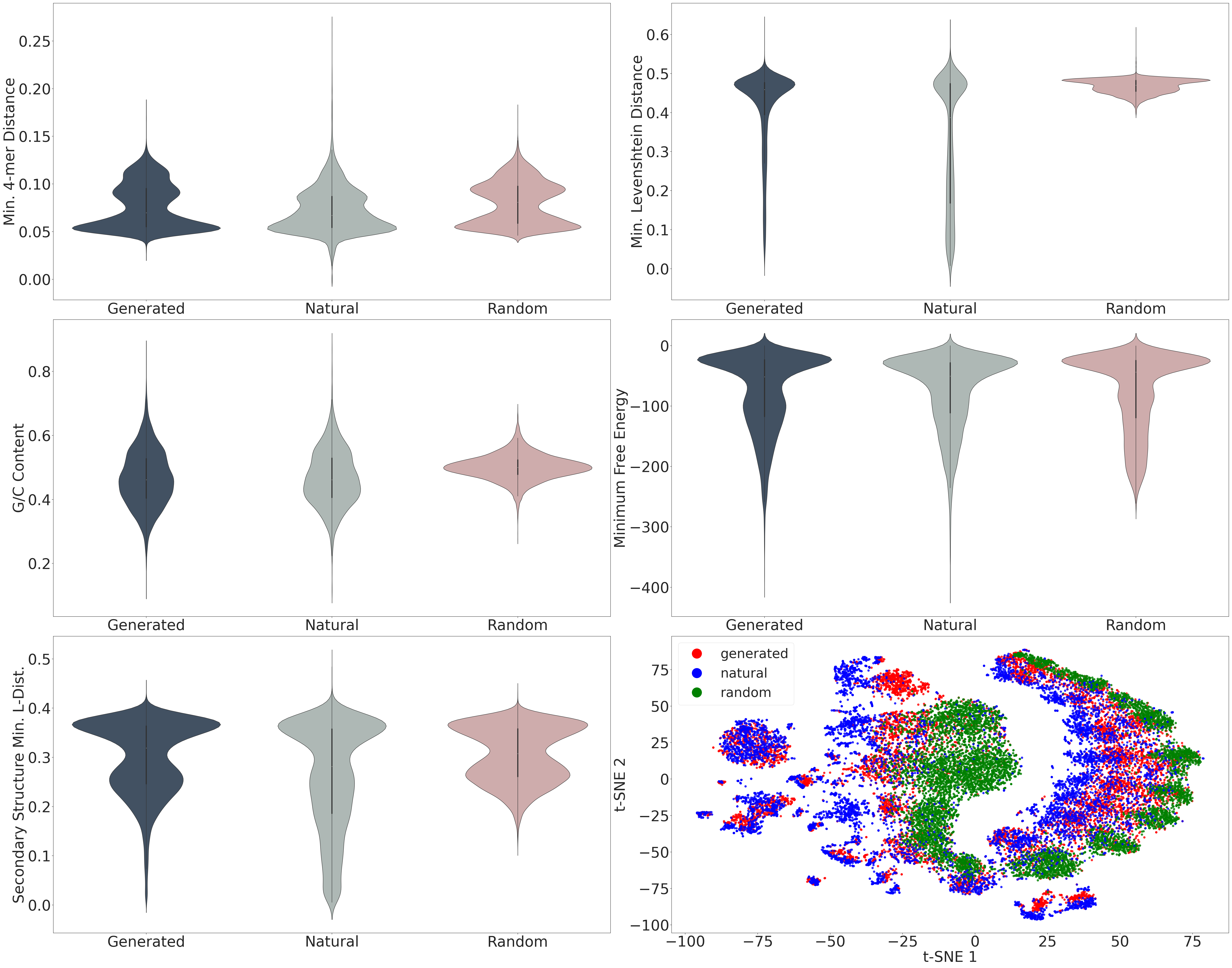}
   \caption{\textbf{Generated samples from \RD compared to natural ncRNAs and random sequences.} Each includes 9000 sequences.  (a) minimum 4-mer distances. (b) minimum sequence Levenshtein distances. (c) G/C content ratios. (d) minimum free energy. (e) minimum Levenshtein distances of RNA secondary structure. (f) t-SNE visualizations of RNAs in latent embedding space. 
   }
   \label{fig:qual}
\end{figure}

Additionally, we create a curated random sequence set by replacing each token in the reference sequences with nucleotide randomly sampled at a probability of 0.5 for G/C, approximating the average G/C content ratio of natural ncRNAs, while maintaining the original sequence length. Next, we plot the distributions of the aforementioned metrics of our generated sequences and compare them to a held-out natural ncRNA reference set and the set of random sequences in Figure~\ref{fig:qual}, which demonstrates that our generated sequences exhibit greater similarity to the natural sequences compared to the random sequences in terms of sequence-level metrics (Levenshtein and 4-mer distances, G/C content ratios), structural-level metrics (Minimum Free Energies and secondary structure distances), and latent space similarities (t-SNE maps).

\section{Optimizing 5'-UTR Sequences with Guided Diffusion}
\label{sec:exp:guidance}

The 5' untranslated region (5'-UTR), located at the beginning of mRNA before the coding sequence, plays a crucial role in regulating protein synthesis by impacting mRNA stability, localization, and translation. 
In this section, we focus on generating 5'-UTR sequences with high protein production levels through guided diffusion. 

We consider two functional properties of 5'-UTRs that correlate with protein production levels~\citep{sample2019human,cao2021high,chen2022interpretable,zheng2023translation,karollus2021predicting}: (1) \textbf{Mean Ribosome Loading (MRL)} measures the number of ribosomes actively translating an mRNA, reflecting its translation efficiency~\citep{sample2019human}. In molecular biology, techniques like polysome profiling (Ribo-seq) are employed to determine MRL. (2) \textbf{Translation efficiency (TE)} represents the rate at which mRNA is translated into protein~\citep{cao2021high}. It is determined by dividing the Ribo-seq RPKM~\citep{ingolia2009genome}, which shows ribosomal footprints, by the RNA-seq RPKM that measures the mRNA's abundance in the cell. This calculation helps gauge how efficiently an mRNA is being translated, independent of its expression level.

\subsection{Fine-tuning \RD for 5'-UTRs} 

\captionsetup[figure]{font=footnotesize,labelfont=footnotesize}
\begin{wrapfigure}[12]{r}{0.29\textwidth}
\centering
\vspace{-15pt}
\includegraphics[width=0.29\textwidth]{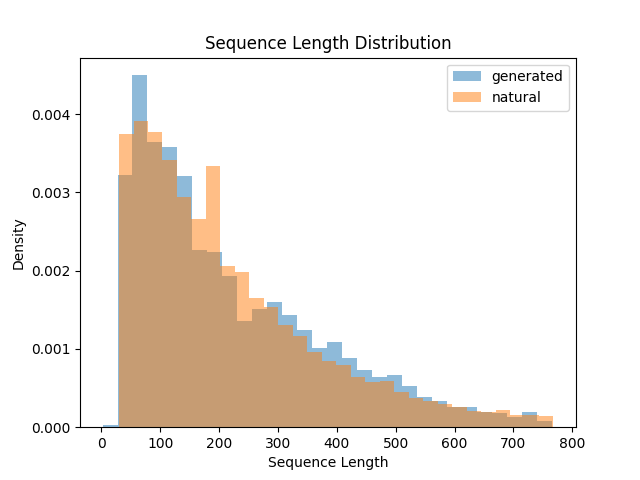}
\vspace{-12pt}
\caption{
\footnotesize{
\textbf{Sequence length histogram} of the natural 
\textbf{UTR} test set and generated sequences (sample size: 20000).
}}
\vspace{-100pt}
\label{fig:seq_len_utr}
\end{wrapfigure}
\captionsetup[figure]{font=normalsize,labelfont=normalsize}

Although both UTRs and ncRNAs share similarities in their non-coding properties, UTRs contain important functional elements that are irrelevant to ncRNAs.
Thus, our pretrained \RD could not be directly applied to generate UTR sequences due to the distributional shift between UTRs and ncRNAs.
To adapt \RD to generate 5'-UTR sequences, we \textit{fine-tune} the score function of \RD using the five-species 5'-UTR dataset from UTR-LM~\citep{chu20245}, where we elected sequences with lengths less than 768 and get 205K samples. We finetune the diffusion model for 1 epoch on top of the frozen sequence auto-encoder with $(K,D)=(32,160)$. 

Upon testing the autoencoder on UTR test set, it shows considerable low reconstruction errors (NLL$=0.0001$, NED=$0.01\% \pm 0.31 \%$), confirming that the autoencoder pretrained with ncRNAs could be transferred onto UTRs.
To compare the generated sequences of the fine-tuned diffusion model with natural 5'-UTRs, we present the sequence length histograms (Figure~\ref{fig:seq_len_utr}) and compute the metrics proposed in Section \ref{sec:exp:eval} (Figure~\ref{fig:qual_utr} in Appendix~\ref{appendix:exp:utr_ft}). The results demonstrate that they align closely with the natural distribution. 

\subsection{Latent Reward Modeling for MRL and TE}  
\label{sec:latent_reward}

\textbf{Datasets.} For the \textit{TE prediction} task, we use three endogenous human 5'-UTR datasets analyzed by \citet{cao2021high}, which contain 28,246 sequences and their measured TE. We reserve 10\% of the dataset for testing, 10\% for validation, and use the remaining 80\% for training.
For the \textit{MRL prediction} task, following the length-based held-out testing approach recommended by \citet{sample2019human}, we train our model on a collection of 83,919 random 5'-UTR sequences ranging from 25 to 100 bp with MRL measurement. We reserve $10\%$ of the data for validation. Subsequently, we test the model on 7,600 human 5'-UTR sequences.

\textbf{Model Architecture.} The latent reward networks take as input a list of $K$ latent vectors obtained by Q-former. We construct a 6-layer 1D convolutional neural network with residual blocks~\citep{he2016deep} and apply it on both the MRL and TE prediction tasks.
Additional details are deferred to Appendix~\ref{appendix:exp:rm}.

\textbf{Results.} For each task, we train two reward models on top of the latent spaces from the pretrained RNA Sequence Autoencorder ($(K,D) = (32,160)$), each initialized under distinct random seeds. One model is utilized for guided generation in Section \ref{sec:guided} while the other is employed for validation assessment. 
We calculate the Spearman 
correlation coefficients between the predicted rewards and the actual rewards (shown in Table~\ref{Table2}).
Notably, the Spearman R values on the held-out test sets are $0.56$ for the TE task and $0.69$ for the MRL task, achieving slightly inferior yet comparable performance compared with baselines~\citep{cao2021high, sample2019human}) and the state-of-the-art predictor built upon UTR-LM (~\cite{chu20245}).

\begin{table*}[ht]
    \centering
    \caption{\textbf{Spearman R-values of MRL and TE reward modeling.} The embeddings after Q-Former remain useful for downstream tasks, allowing us to build effective reward models based on them.   
    }
    \resizebox{0.7\linewidth}{!}
    {
      \begin{tabular}{l||c|c|c||c|c|c}
      \toprule 
& \multicolumn{3}{c||}{TE} &  \multicolumn{3}{c}{MRL} \\
          &  Ours  & UTR-LM~\citep{chu20245} & \citep{cao2021high} &  Ours & UTR-LM~\citep{chu20245} &  \citep{sample2019human} \\
        \midrule
      Test Spearman R &  0.56 & \textbf{0.64}  & 0.60  & 0.69  & \textbf{0.85} & 0.83  \\
        \bottomrule
        \end{tabular}
      }
    \label{Table2}
 \end{table*}%
 
\subsection{Reward-Guided Diffusion}
\label{sec:guided}
We compute gradient-based forward universal guidance~\citep{bansal2023universal} in the latent space with the trained reward models to shift the latent backward sample paths towards generating high MRLs/TEs sequences as described in Section~\ref{sec:method:dm:guidance}, where we choose the loss function $\ell$ to be $L_2$ loss and vary the target reward $r^*$ and the guidance strength $\lambda$.
For guidance, we take the latent reward models trained with one random seed for TE and MRL optimization. 
For cross-validation, we employ the reward models trained with another random seed as the \textit{validation} model to evaluate the rewards of the generated sequence.
The best performing guidance parameters are able to improve the mean TE by up to $166.7\%$ and mean MRL by up to $52.6\%$ compared to the unguided baseline. We defer our systematic analysis of the effects of different guidance parameters to Appendix~\ref{sec:ablation:guidance:parameters}. 

\textbf{Baselines.} We compare the performance of our guidance pipeline with the following two baselines:
\begin{itemize}[itemsep=1pt, parsep=1pt, topsep=2pt]
    \item \textbf{Random}. We adopt a best-of-$N$ sequence generation method similar to Sample et al.~\cite{sample2019human}, which uses a genetic algorithm to generate random sequences and merely retain ones with high rewards. In our approach, we generate random sequences as in Section ~\ref{sec:results_nc} and select the sequence with the highest reward according to our guidance reward model.
\item \textbf{UTRGAN}, a 5'-UTR sequence generative model targeting translation rate optimization. We directly generate 6000 reward-optimized sequences using the official codes release by UTRGAN~\cite{barazandeh2023utrgan}.
\end{itemize}

\textbf{Results.} For each generated sequence, we evaluate the reward using the \textit{validation} reward models and also calculate the Minimum Free Energy (MFE) as the proxy for structural stability. 
We plot the validation reward (MRL/TE) v.s. MFE of samples generated by different methods in the Pareto fronts (Figure~\ref{fig:pareto}). 
Our gradient-guided latent diffusion model achieves higher rewards with the same MFE level compared to the two baselines. This demonstrates that our method has a better reward–structural stability \textit{trade-off}.
\input{figs/pareto}

%% file: figs/pareto.tex
\begin{figure}[h] 
\vspace{-2mm}
    \centering
\subfigure[MRL]
{
\includegraphics[width=0.4\textwidth]{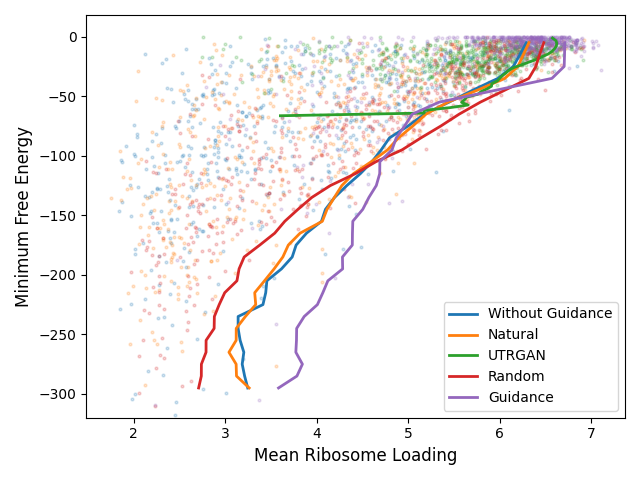}
} 
\subfigure[TE]
{
\includegraphics[width=0.4\textwidth]{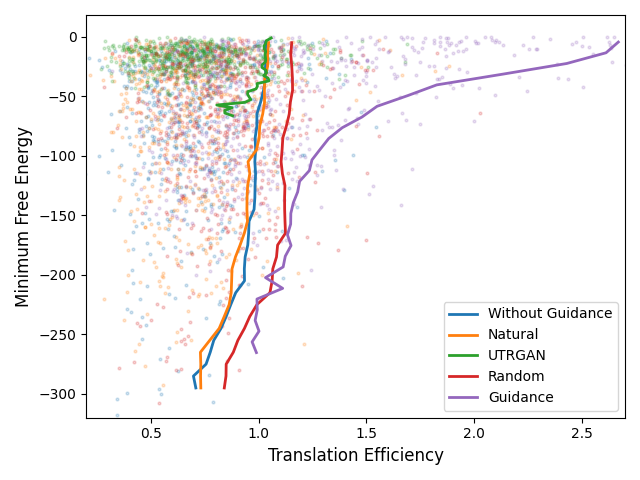}
}
\caption{\textbf{Pareto front curves} between Minimum Free Energy (MFE) and Mean Ribosome Loading (MRL)/Translation Efficiency (TE). The curves are generated by selecting the top 10\% quantile MRL/TE within sliding windows around each MFE value. Each shaded dot denotes a sequence. All the sequences are evaluated with the same \textit{validation} reward models.
}
\label{fig:pareto}
\vspace{-1em}
\end{figure}

%% file: sections/related_work.tex
\section{Related Work}
\label{sec:related:work}

People have been exploring building D3PM-type~\citep{austin2021structured} discrete diffusion models and Diffusion-LM type~\citep{li2022diffusion} for proteins sequences and 3D structures~\citep{lee2023score, wu2024protein, gao2023diffsds, anand2022protein,lin2023generating, trippe2022diffusion, Watson2022.12.09.519842, yim2023se, ingraham2023illuminating, gruver2023protein, alamdari2023protein,Luo2022.07.10.499510} and DNAs~\citep{avdeyev2023dirichlet, li2024discdiff,ferreira2024dna}. 
Besides, latent diffusion models for text sequences have been studied in~\citep{xu2022versatile, zhang2023planner, lovelace2022latent}. 
For a detailed discussion on continuous/discrete diffusion models and their applications in biology, see the full related work section in Appendix~\ref{appendix:related:work}.

%% file: sections/conclusion.tex
\section{Conclusion}
We present \RD, a diffusion-based generative model that captures the RNA sequence distribution within a latent space constructed by a Sequence AutoEncoder, which utilizes a Querying Transformer to manage the variable-length nature of RNA sequences. By incorporating gradient guidance from latent reward models, our diffusion model can generate novel 5'-UTRs with up to $166.7\%$ higher TE and $52.6\%$ higher MRL, while achieving a superior \textit{trade-off} between rewards and structural stability. These results provide a biologically relevant demonstration of RNA sequence generation, an underexplored yet exciting area of research.

%% file: sections/acknowledgement.tex
\section*{Acknowledgments}
We thank Zaixi Zhang for valuable suggestions to the paper.

%% file: appendix/limitation.tex
\section{Limitations}
\label{appendix:limitation}

The performance of our latent diffusion models depends on the total dimension of the latent space $K \times D$, where $K$ is the number of query embeddings and $D$ is the embedding dimension for each query. Increasing $K$ and $D$ makes the reconstruction task and the reward prediction tasks easier, but it allows more redundant information in the latent space, which may hinder the training of the diffusion model. Therefore, we require a hyperparameter search to find the best-performing $K$ and $D$.
To perform reward-guided diffusion, we require a trained differentiable reward network, whose performance crucially depends on the number and the quality of the labeled data. 
To extend to non-differentiable rewards, additional fine-tuning with reinforcement learning methods~\citep{black2023training} is still required. 
Besides, we did not synthesize the generated 5'-UTRs in wet lab experiments to test the performance, which we leave as future work.

%% file: appendix/impact.tex
\section{Broader Impacts}
\label{appendix:impact}

Our work aims to generate 5'-UTRs with optimized protein production levels, which could significantly enhance the understanding and engineering of gene expression. This can lead to breakthroughs in developing new therapies for diseases such as cancer and genetic disorders.
It can also improve synthetic biology systems that produce valuable compounds, such as pharmaceuticals and bio-fuels, more efficiently and sustainably.
Nevertheless, the capability to alter gene expression and produce biological molecules could be misused to create harmful biological agents or cause unintended harm to ecosystems.

%% file: appendix/related_work.tex
\section{Additional Related Work}
\label{appendix:related:work}

\subsection{Continuous Diffusion Models} 
A majority of works on diffusion models focus on modelling continuous data~\citep{ho2020denoising, song2020denoising, ramesh2022hierarchical, rombach2022high, balaji2022ediffi, nichol2022point, ho2022imagen, ajay2023is, janner2022diffuser, liang2023adaptdiffuser, tashiro2021csdi, Watson2022.12.09.519842}. During training, continuous Gaussian noises are sequentially injected to the sample. The core of diffusion models is a denoising network that takes in a Gaussian-noise-perturbed sample and predicts the clean version of the sample. Backed by the theory of forward/backward SDEs \citep{song2020score, oko2023diffusion, lee2023convergence, chen2022sampling, chen2023score, anderson1982reverse, haussmann1986time}, diffusion models are guaranteed to recover the true data distribution.

The behavior of diffusion models can be controlled through \textit{guidance}, where an additional guidance term is added into each step of the backward diffusion process to drive the sample toward the regions with desired properties~\citep{dhariwal2021diffusion, bansal2023universal, chung2022diffusion, graikos2022diffusion,  guo2024gradient, yuan2024reward}. Classifier-based guidance~\citep{dhariwal2021diffusion} and its extentions utilize a separate classifier or a reward model to calculate the guidance term, while classifier-free guidance~\citep{ho2022classifier} calculates the guidance term based on the conditional diffsuion model itself. More complated controls can be done via finetuning the diffusion models with additional networks~\citep{zhang2023adding}.

\subsection{Diffusion Models for discrete variables}
\label{sec:related:discrete} 

The essential components of diffusion models are the forward process, which defines how a clean sample is sequentially corrupted into a pure random noise, and the corresponding backward process, which involves training a denoising network to learn the score function to revert the forward process and sequentially recover the input.

\textbf{Diffusion on Discrete Domains.} To extend diffusion models to discrete domains, the forward/backward proocesses and the score matching objectives need to be redefined~\citep{austin2021structured, johnson2021beyond, savinov2021step, hoogeboom2021autoregressive}. Notably, BERT-type encoder-only models can be naturally modified into diffusion models, where masking \& demasking operations serve as the noising/denoising counterparts~\citep{austin2021structured, he2022diffusionbert, ye2023diffusion}. Researchers have also explored score matching for discrete variables~\citep{meng2022concrete, sun2022score, campbell2022continuous, lou2023discrete}. 

\textbf{Diffusion on Continuous Embedding Spaces}. One straightforward idea to circumvent the discreteness is to learn diffusion models on the continuous embedding spaces. The pioneering work Diffusion-LM~\citep{li2022diffusion} proposes to jointly learn a token embedding together with the continuous diffusion model and develops the re-parametrization method and the clamping trick to reduce rounding error induced when projecting the embeddings back to discrete tokens. Afterwards, various work have been done to extend the Diffusion-LM framework to conditional generation and improve the token embeddings and training~\citep{gong2022diffuseq, lin2023text, gao2022difformer, chen2022analog, han2022ssd, han2023ssd, avdeyev2023dirichlet, strudel2022self, dieleman2022continuous, gulrajani2024likelihood}.

\textbf{Model Structures.}
As BERT-type models~\citep{devlin-etal-2019-bert, liu2019roberta} have demonstrated success in reconstructing the masked tokens from the remaining tokens, people adapt Transformers as the denoising networks for sequences, either operating directly on the discrete space with adding masks as one type of sample corruption, or on the continuous embedding space with the standard Gaussian noise corrupting.

\subsection{Latent Diffusion Model for Sequences}  
\label{appendix:related:work:ldm}

Latent diffusion models for sequences are essentially sequence variational auto-encoders with diffusion latent priors, which contain a chain of conditional Guassian distributions, rather than a simple Gaussian distribution. The difference between latent diffusion models and the Diffusion-LM scheme~\citep{li2022diffusion} (Section~\ref{sec:related:discrete}) lies in that latent diffusion models incorporate complex sequence (variational) auto-encoder models while Diffusion-LM only utilizes a token embedding layer.

\textbf{Sequence Variational Auto-encoders.} The design of sequence VAEs can be traced back to the era of RNNs~\citep{bowman-etal-2016-generating, kim2018semi, he2019lagging}. Afterwards, OPTIMUS~\citep{li-etal-2020-optimus} is a sequence VAE that connects pretrained BERT and GPT-2, where the authors only use the \verb|<cls>| token embedding of BERT to extract the latent vector. Similarly, \citet{park-lee-2021-finetuning} finetune a pretrained T5 model ino a sequence VAE and they use mean/max pooling over all the embeddings to obtain the latent.

A notorious challenge for sequence VAEs (with Gaussian priors) is the KL vanishing problem~\citep{bowman-etal-2016-generating}, where the encoder produces nearly identical latents for all the inputs (a.k.a. posterior-collaspe) and the decoder completely ignores the latents. The intuitive reason behind the KL vanishing problem is that the KL-regularization is too strong and the encoder fails to extract useful information for the decoder to reconstruct the sequence. Several tricks are proposed to address the issues, such as anealling the KL-regularization coefficient, memory dropout, etc~\citep{li-etal-2020-optimus, yang2023dior}.

\textbf{Latent Diffusion Model for Sequences.} People have been building latent diffusion model for sequences, such as Versatile Diffusion~\citep{xu2022versatile}, LD4LG~\citep{lovelace2022latent}, Planner~\citep{zhang2023planner}.
With diffusion priors, the KL vanishing problem can be reduced by using a tiny or zero KL-regularization, as the diffusion models are strong enough for modeling complex multimodal distributions~\citep{rombach2022high}.
Versatile Diffusion~\citep{xu2022versatile} trains a textual diffusion model on top of the latent space of OPTIMUS, with a focus on the multimodality setting, e.g. image captions.

\textbf{Addressing the similarities and differences of Planner~\citep{zhang2023planner}.}

Planner~\citep{zhang2023planner} proposes to use the first $K$ token embeddings from the BERT model as the latent variable to increase the information within the latent embedding, but the BERT model needs to be fine-tuned. 
In comparsion, we keep the pretrained encoder model fixed to avoid the deterioration of the encoder, e.g., the encoder may be finetuned to forget the semantics, as the reconstruction task only requires the sequence auto-encoder to reproduce the raw input sequence.
We train a Q-Former~\citep{li2023blip} to better summarize the entire embedding sequence through cross-attention.  
Our experiment on MRL/TE predictions also empirically verifies that the latent vectors contain biologically meaningful information.

\textbf{Addressing the similarities and differences of LD4LG~\citep{lovelace2022latent}.}

In methodology, the most similar work to us is LD4LG~\citep{lovelace2022latent}. Specifically, LD4LG uses a pretrained encoder-decoder model such as BART~\citep{lewis-etal-2020-bart} or T5~\citep{JMLR:v21:20-074}. Similarly the encoder is fixed and a Perceiver Resampler~\citep{alayrac2022flamingo} is trained to summarize the embeddings. Perceiver Resampler and Q-Former are essentially the same in this setting. 

However, LD4LG incorporates an additional transformer Reconstruction Network, which transforms the latent embeddings for the fixed pretrained decoder to cross-attend to. 
Instead, we perform a simple linear transform to the latent vectors and use them as the soft prefix for the decoder. We initialize the decoder from stratch and train the decoder together with the Q-Former, so the decoder focuses only on translating the latent variables back to discrete sequences. Our design choice reduces the computation for the pipeline and also does not require a paired pretrained encoder-decoder model to work.
Additionally, LD4LG solely focuses on NLP tasks such as summarization and translation. Our work focuses on RNA sequence modeling. We further show that our latent embeddings can be used to predict reward of interests (MRL and TE), and study using the trained reward model as guidance to optimize the RNA sequence.

\subsection{Generaitve Models for Biological Sequences}  

{Generative models have achieved considerable success in designing biological sequences~\citep{guo2024diffusion}, with applications ranging across protein and DNA. Among these, diffusion models have emerged as a particularly powerful tool for protein design, enabling the generation of large and diverse protein structures through guided iterative processes. This overcomes the limitations of traditional deep generative models like VAEs and GANs~\citep{anand2018generative,lin2021deep,eguchi2022ig,anishchenko2021novo,greener2018design,anand2019fully,karimi2020novo,Sevgen2023.01.23.525232}, which often generate only small proteins or protein domains.} {For instance, ProteinSGM~\citep{lee2023score} employs diffusion models to generate protein structures from residue distances and angles, enabling both unconditional and targeted generation with post-processing by Rosetta~\citep{simons1999ab}. Foldingdiff~\citep{wu2024protein} uses a transformer model and DDPM to unconditionally model protein backbones using angles, while DiffSDS~\citep{gao2023diffsds} enhances backbone structure quality by using an encoder-decoder language model to generate a 1D directional representation of invariant atom features for the diffusion process. The SE(3)-equivariant~\citep{thomas2018tensor,satorras2021n} DDPMs~\citep{anand2022protein} generate fully atomistic protein structures and sequences by equipping with invariant point attention~\citep{jumper2021highly} structural modules. Genie~\citep{lin2023generating} improves noise prediction with geometric asymmetry in residue coordinates. SMCDiff~\citep{trippe2022diffusion} utilizes a GNN~\citep{scarselli2008graph} within a diffusion model to address motif-scaffolding in proteins, utilizing conditional sampling formulated as a sequential Monte Carlo simulation solved by particle filtering. RFdiffusion~\citep{Watson2022.12.09.519842} combines DDPM with RoseTTAFold for generating unconditional large protein generation and conditional scaffolding using a self-conditioning strategy. FrameDiff~\citep{yim2023se} efficiently generates protein backbones using SE(3)-diffusion with denoising score matching. Chroma~\citep{ingraham2023illuminating} uses a GNN-based diffusion model to create large, diverse proteins and protein complexes conditioned on specific properties.}

{Guided diffusion models significantly enhance protein design by integrating generative and discriminative approaches to optimize sequence fitness. Diffusion Optimized Sampling (NOS) \citep{gruver2023protein} boosts diffusion models with gradient-guided denoising and incorporates LaMBO, a Bayesian optimization process that accommodates multiple objectives and edit-based constraints. The EvoDiff framework \citep{alamdari2023protein} utilizes evolution-guided generation to create proteins that are structurally plausible and meet evolutionary and functional requirements, pushing protein engineering towards a programmable, sequence-first paradigm. Additionally, DiffAb \citep{Luo2022.07.10.499510} applies an antigen-specific guided diffusion method to iteratively refine antibody CDR sequences and structures based on the 3D structure of targeted antigens, facilitating the precise and efficient design of functionally optimized antibodies.} 
 
{In the realm of DNA, the application of diffusion models is emerging but less extensive than in protein design. The Dirichlet Diffusion Score Model~\citep{avdeyev2023dirichlet} specifically generates human promoter DNA sequences, exploring structural plausibility and diversity. Alongside, latent diffusion models such as DiscDiff~\citep{li2024discdiff} and DNA-Diffusion~\citep{ferreira2024dna} have tailored the diffusion approach for discrete data, generating DNA sequences and regulatory elements that closely mimic natural genomic variability. These models transform discrete data into a continuous latent space, a necessary adaptation for applying diffusion principles that require differentiable probability density functions.}

{Despite the successes with proteins and DNA, the application of diffusion models to RNA, particularly 5'-UTR sequences, has been limited. UTRGAN~\citep{barazandeh2023utrgan} uses GANs to generate novel 5'-UTR sequences, establishing itself as the state-of-the-art generative model for this purpose. In this study, we focus on generating ncRNAs with latent diffusion models and adopt guided diffusion for high MRL/TE 5'-UTR generation as a downstream task.}

%% file: appendix/exp.tex
\section{Additional Experimental Results}

\subsection{Additional Experimental Setup of RNA AutoEncoder}
\label{appendix:exp:ae}

\textbf{Details of the Models of Sequence AutoEncoder.} We adopt the pretrained RNA-FM~\citep{chen2022interpretable} as the encoder. RNA-FM is an encoder-only model trained with masked language modelling on non-coding RNAs (ncRNAs), which can extract biologically meaningful embeddings from ncRNAs and can be adapted for various downstream tasks, such as UTR function prediction, RNA-protein interaction prediction, secondary structure prediction, and etc. 
For the Q-Former, we modify the ESM-2 model architecture~\citep{lin2022language} by adding $K$ trainable query token embeddings and adding cross-attention layers in each transformer block that attends to the embeddings from the RNA-FM. 
For the decoder model, we adopt the same model architecture of ProGen2-small~\citep{nijkamp2023progen2} with our customized tokenizers for RNAs.

\textbf{Notes on ESM-2 and ProGen2.} ESM-2~\citep{lin2022language} is an encoder-only model trained on protein sequences. The model architecture of ESM-2 improves upon BERT by replacing the absolute positional embeddings with Rotary Position Embedding (RoPE)~\citep{su2024roformer}. The model architecture of ESM-2 has been adapted to model other biological sequences as in RNA-FM~\citep{chen2022interpretable}, UTR-LM~\citep{chu20245}, etc. 

ProGen2 models~\citep{nijkamp2023progen2} are originally designed for protein sequence modeling. The architecture differs from standard GPT-like models in (1) Rotary Position Embedding and (2) parallel feed-forward and attention computation. 

\textbf{Preprocessing.}
We adopt the following data preprocessing method following~\citet{chen2022interpretable}. First, we replace base T with base U. Next, we eliminate identical sequences by applying the CD-HIT-EST clustering algorithm with a cut-off at 100\%~\citep{fu2012cd}. Afterwards, we remove RNA sequences with lengths $> 768$ and RNA sequences containing rare symbols other than A,U,G,C. The resulting dataset contains 1.1M RNA sequences.
We adopt a simple tokenization method, where each base (A,U,G,C) is viewed as one token.

\textbf{Training details.} 
We split the dataset into the train split and the test split using a 9:1 ratio. 
We tune the number of query tokens $K$ among $\{16, 32\}$ and the hidden dimension $D$ of each latent vector among $\{40, 80, 160, 320\}$. 
We freeze the encoder and pretrain the Q-Former and the decoder from scratch using Adam optimizer with initial learning rate $1e{-4}$ and the cosine learning rate schedule. 
For $(K, D) = (16, 160), (16, 320), (32, 320)$,
we pre-train the auto-encoder for 1 epoch; while for 
$(K, D) = (32, 160)$,
the pre-training processes last 2 epochs 
and the remaining settings take 3 epochs to converge.

\textbf{Experiments Compute Resources} We train the auto-encoder on one NVIDIA H100-80G GPU. The estimated time for one epoch of training is around 8.5 hours.

\subsection{Additional Experimental Setup of Latent Diffusion Models}
\label{appendix:exp:dm}

\textbf{Models.} We adopt the following modifications to our Transformer denoising networks: (1) we add a skip-connection directly between the final output and the input, and (2) we use linear projection layers before and after the transformer to accommodate for different latent dimensions.

\textbf{Training Details.} We train a diffusion model on top of the latent space of each of the auto-encoders we obtain in Section~\ref{sec:exp:autoencoder}. We use $\epsilon$-prediction scheme so the denoising network predicts the added noise of the noisy sample. We adopt the linear noise schedule as DDPM~\citep{ho2020denoising} and set the number of diffusion steps $T=1000$. For each auto-encoder setting, the diffusion model is trained for 3 epochs by using AdamW optimizer with initial learning rate $1e{-4}$ and the cosine learning rate schedule. 

\textbf{Generation Details.} We use the DDIM sampling  method~\citep{song2020denoising} with $\eta=0$ (deterministic) and $50$ denoising steps to get samples from our latent diffusion models. Then we transform the latent vectors into soft prompts for the decoder and auto-regressively generate a sample sequence nucleus-by-nucleus with top-$p=0.95$ and temperature $1$.  
We remove any invalid sample sequence containing special tokens in the middle (e.g., \verb|<bos>|).

\textbf{Experiments Compute Resources} We train the diffusion model on one NVIDIA H100-80G GPU. The estimated time for one epoch of pre-training on 1.1M ncRNA dataset is 3.5 hours; while for one-epoch fine-tuning on 205K 5'UTR dataset, the training process approximately costs 40 minutes.  

\subsection{Additional Experimental Setup of Latent Reward Models}
\label{appendix:exp:rm}

\textbf{Additional Details on Datasets.} 
For the \textit{TE prediction} task, the three endogenous human 5' UTR datasets we used are derived from different cell lines or tissue types: human embryonic kidney 293T (HEK), human prostate cancer cell line PC3 (PC3), and human muscle tissue (Muscle). These datasets contained 14,410, 12,579, and 1,257 sequences, respectively. In training the reward model, we combined these three datasets into a larger one.

\textbf{Model Architecture.} We build Convolutional ResNets on top of the latent space to predict the rewards. Each layer in this network, referred to as a ResBlock, consists of two 1D convolutional layers activated by ReLU functions and followed by Batch Normalization. Convolution operations are applied along the query dimension, with the number of input channels matching the embedding dimension of the output from the querying transformer, and the channel count doubles between ResBlocks. We employ the smallest kernel size of 3 to accommodate our constrained query length, along with padding of 1 to preserve dimensional integrity. The output from the ResNet is averaged across the query dimension before being processed through a Multi-Layer Perceptron (MLP) network to produce the final scalar output.
To enhance model robustness, we incorporate a dropout layer after each Resblock to prevent overfitting --- a strategy we found more beneficial than downsizing the model, which could compromise its learning capacity.

\textbf{Training Details.} We trained the reward model for 50 epochs on the MRL prediction dataset and for 100 epochs on the TE datasets. We utilized the AdamW optimizer, setting the initial learning rate 1e-4 with a warm-up phase of 3,000 steps, followed by a cosine learning rate schedule. We implemented a dropout rate of 0.2 and an Adam weight decay of 0.02 to combat overfitting. We applied data augmentation to further ensure robust learning by introducing random Gaussian noise to the input latent vectors. We adopted the same noise scheduler as used in diffusion models. Instead of sampling noise from all 1,000 steps, we restricted sampling to only the first 10 steps.

\textbf{Experiments Compute Resources} We train the reward models on one NVIDIA A100-80G GPU. The estimated time for training one reward model is 5 hours for the MRL task (50 epochs of 83k samples ) and 2 hours for the TE task (100 epochs of 28k sequences).
\subsection{Ablations Studies on RNA-FM}
\label{appendix:exp:ab_rnafm}
We validate the necessity of the pre-trained RNA-FM in this section. When we replace the pre-trained encoder RNA-FM with naive one-hot embeddings, both the reconstruction capability of the sequence autoencoder and the reward model performance are affected, where we observe significant increases in the reconstruction error and declines in the R-values of the reward model (Table~\ref{tab:ab_rnafm}).

\begin{table}[h]
    \centering
    \caption{\small{\textbf{Ablations on RNA-FM}. We replace RNA-FM with naive one-hot token embedding and test the performances on sequence autoencoder and MRL reward model training under different Q-former settings ($K$, $D$). NLL: reconstruction error, NED: normalized edit distance.}
    }
    \subfigure[($K$, $D$) = (32, 160).]{
        \centering
        \resizebox{0.7\linewidth}{!}{
      \begin{tabular}{r||cc|cc}
      \toprule 
      \multirow{2}{*}{} & \multicolumn{2}{c|}{Sequence AutoEncoder} &  \multicolumn{2}{c}{MRL Reward Model} \\
        & NLL $\downarrow$ &  NED $\downarrow$ & Test Spearman R $\uparrow$ & Test Pearson R $\uparrow$ \\
        \midrule  
      w/o RNA-FM & 0.0018  &  0.10 \% $\pm$ 4.91 \%  & 0.5983 & 0.6969 \\  
      w/ RNA-FM & \textbf{0.0003} & \textbf{0.02\% $\pm$ 0.40 \%} &  \textbf{0.6901} & \textbf{0.7655} \\  
      \bottomrule 
        \end{tabular}
        \label{tab:sub1}
    }
    }
    \hfill
    \subfigure[($K$, $D$) = (16, 160).]{
        \centering
        \resizebox{0.7\linewidth}{!}{
      \begin{tabular}{r||cc|cc}
      \toprule 
      \multirow{2}{*}{} & \multicolumn{2}{c|}{Sequence AutoEncoder} &  \multicolumn{2}{c}{MRL Reward Model} \\
        & NLL $\downarrow$ &  NED $\downarrow$ & Test Spearman R $\uparrow$ & Test Pearson R $\uparrow$ \\
        \midrule  
      w/o RNA-FM & 0.1755  &  30.30 \% $\pm$ 36.13 \%  & 0.2526 & 0.3377 \\  
      w/ RNA-FM & \textbf{0.0156} & \textbf{2.19\% $\pm$ 5.50 \%} &  \textbf{0.3699} & \textbf{0.4599} \\  
      \bottomrule 
        \end{tabular}
        \label{tab:sub2}
    }
    }    
    \label{tab:ab_rnafm}% 
\end{table}%

\clearpage
\subsection{Evaluation of 5'-UTR \RD Fine-tuning}
\label{appendix:exp:utr_ft}
\begin{figure}[h]
  \centering
  \includegraphics[width=1\linewidth]{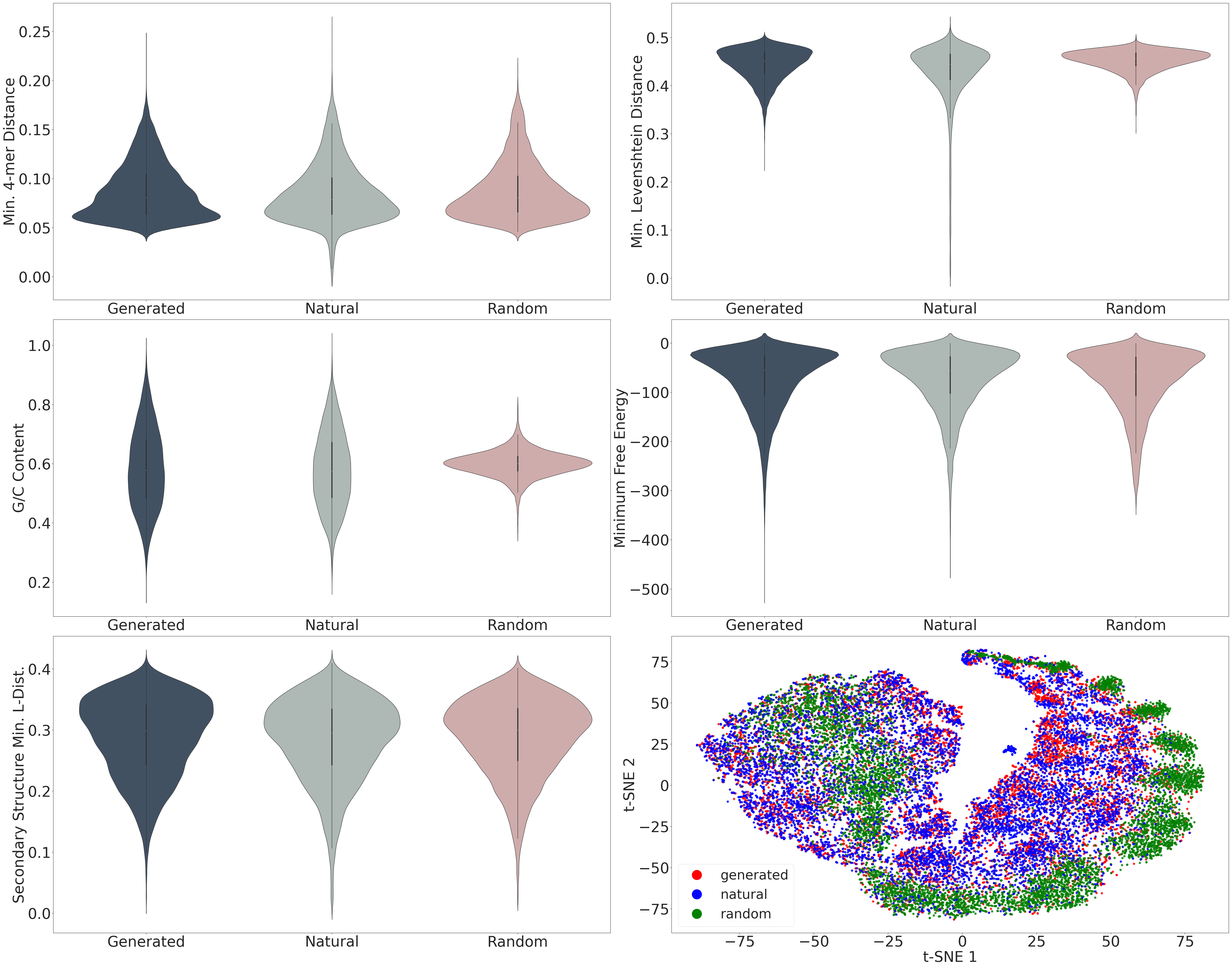}
   \caption{Comparing the sequences generated by the \emph{fine-tuned} latent diffusion model with natural 5'UTR sequences (test set) and random sequences in terms of biological metrics. (a) minimum 4-mer distances. (b) minimum sequence Levenshtein distances. (c) G/C content ratios. (d) minimum free energy. (e) minimum second structure Levenshtein distances. (f) t-SNE of latent space. The random sequences follow the same variable length distribution as natural UTR sequences, and each nucleotide is sampled at a G/C content ratio of 0.6, resembling the average level of natural UTR. 
   }
   \label{fig:qual_utr}
\end{figure}

We compare the samples generated by fine-tuned latent diffusion models with natural 5'UTR reference dataset and random samples following the same evaluation procedure in Section \ref{sec:exp:eval}. The results are shown in Figure \ref{fig:qual_utr}.

\clearpage
\subsection{Ablations on Guidance Parameters for Guided 5'UTR Generation}
\label{sec:ablation:guidance:parameters}
\begin{figure}[!htb]
\centering
\subfigure[TE Optimization]{\includegraphics[width=0.75\textwidth]{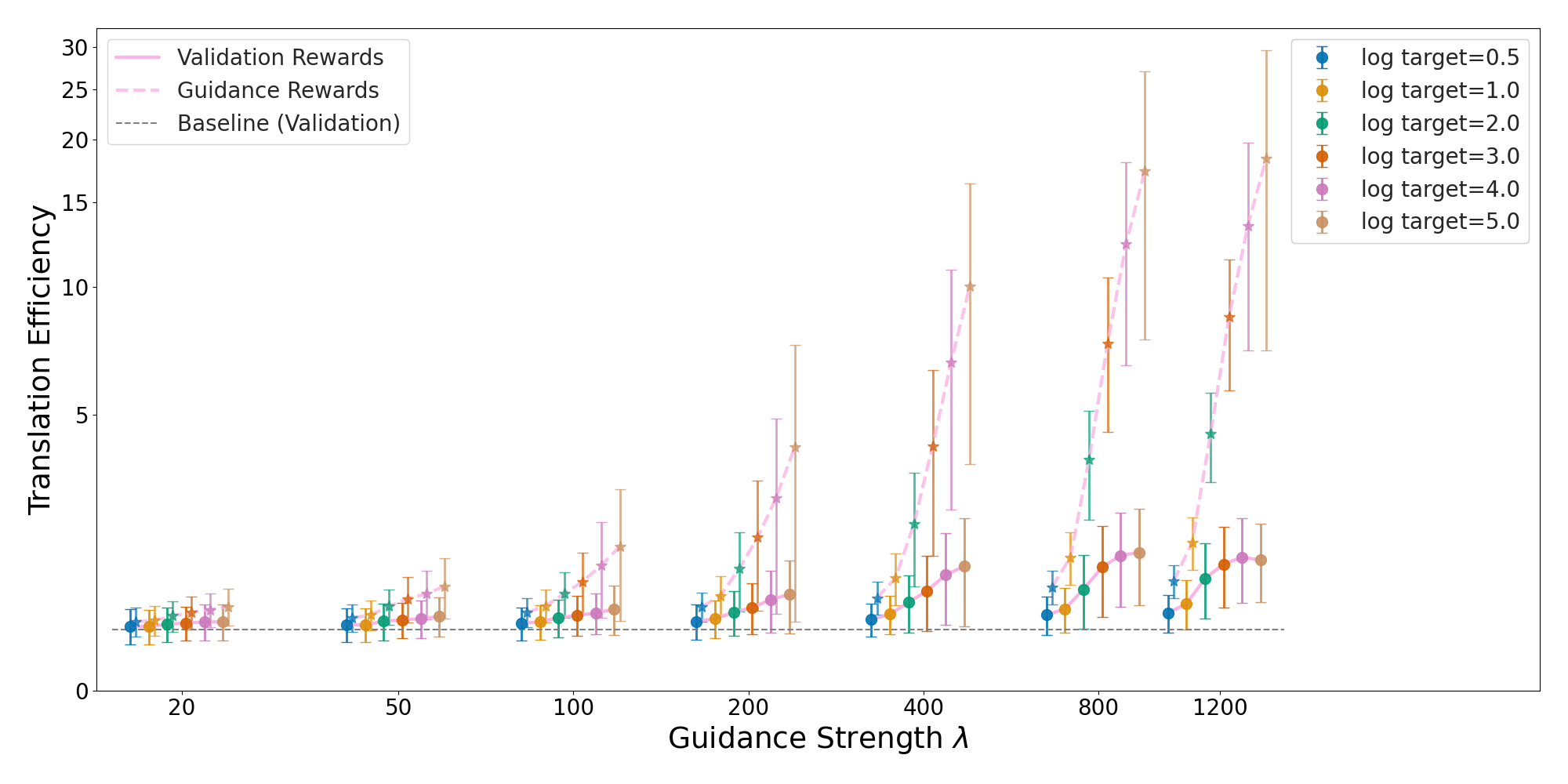}
}
\subfigure[MRL Optimization]{\includegraphics[width=0.75\textwidth]{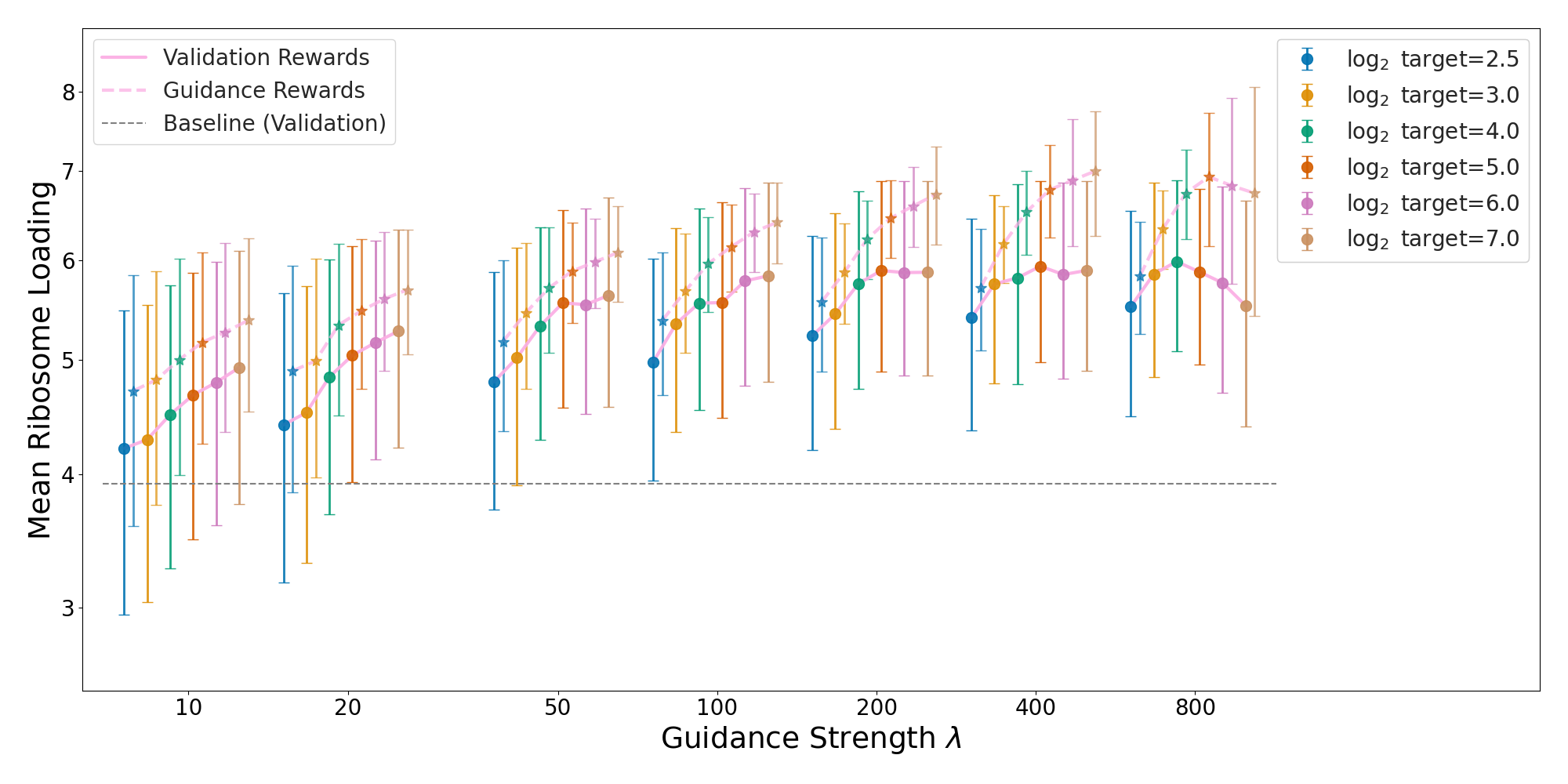}}
\caption{\textbf{Performances of reward-guided diffusion on TE and MRL tasks.} Baseline results refer to the evaluation of unguided generation with validation reward models. Guidance Reward Evaluation indicates the assessment with the same reward models which are used for guided generation. 
All evaluated sequences are over 30-bp. 
} 
\label{fig:reward_optim}
\end{figure}

\textbf{Guidance Parameter Search.} We tune the guidance for generation by varying the values of guidance strength $\lambda$ and target reward $r^*$ as of Equation \ref{eq:pred_guidance}, and plot the rewards of guided generation under different guidance parameters $\lambda$ and $r^*$ in Figure~\ref{fig:reward_optim}. Among all the guidance settings, the best \textit{validation} TE 
($1.84\pm0.82$)
is achieved with the moderate guidance strength as 
800
and the {log}-target value as 5 
, showing 
$\textbf{166.7\%}$
improvement compared to the \textit{validation} TE of unguided generation 
($0.69\pm0.22$).
Under the MRL-guidance setting of 
$\lambda=800$ and $\log r^*=4$,
mean \textit{validation} MRL of the generated sequences increases from 
$3.92\pm1.31$ to $5.98\pm0.90$ with maximum relative increase of $\textbf{52.6
\%}$.

\begin{figure}[!htb]
\centering
\subfigure[G/C Content]{\includegraphics[width=0.98\textwidth]{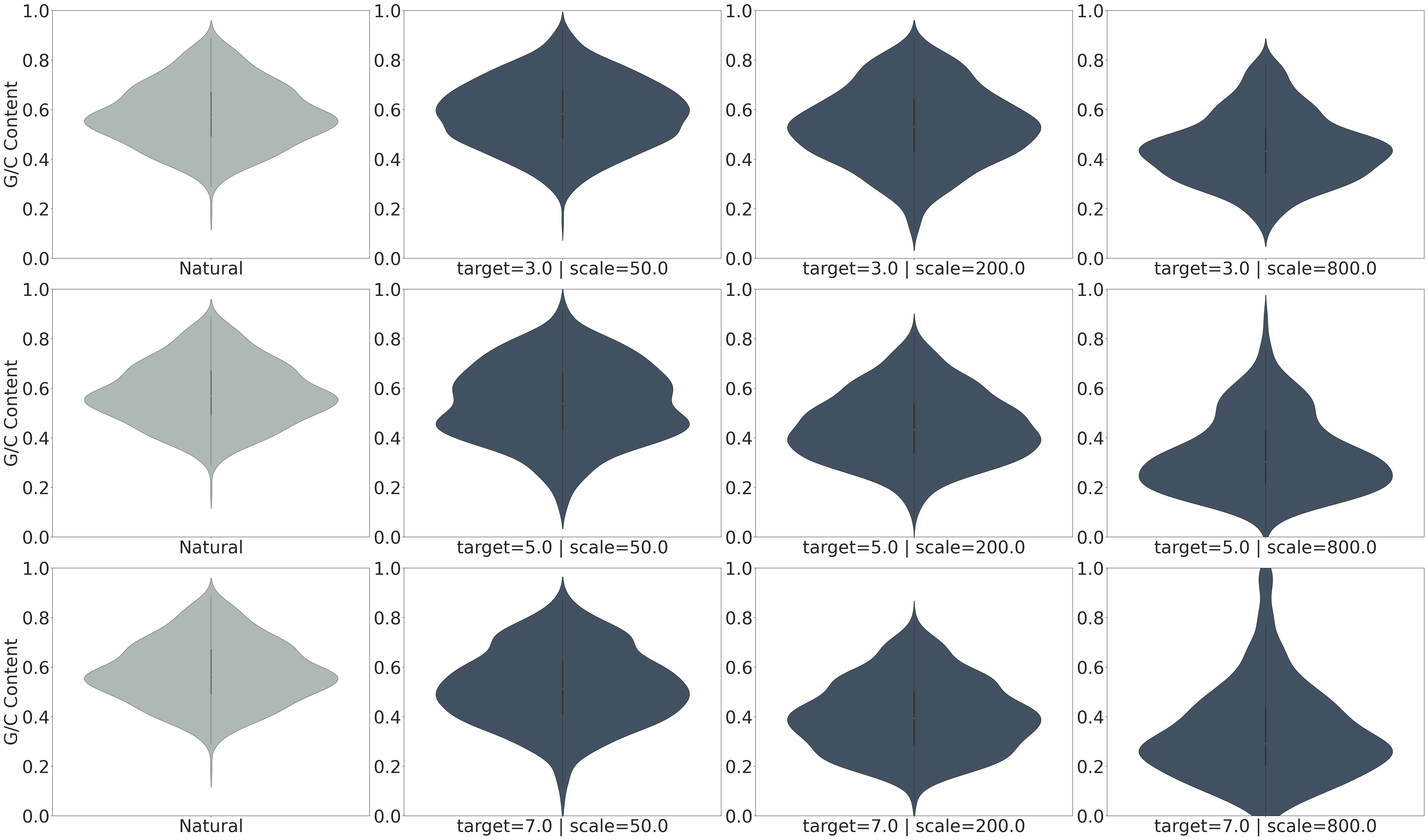}}
\subfigure[Minimum Free Energy]{\includegraphics[width=0.95\textwidth]{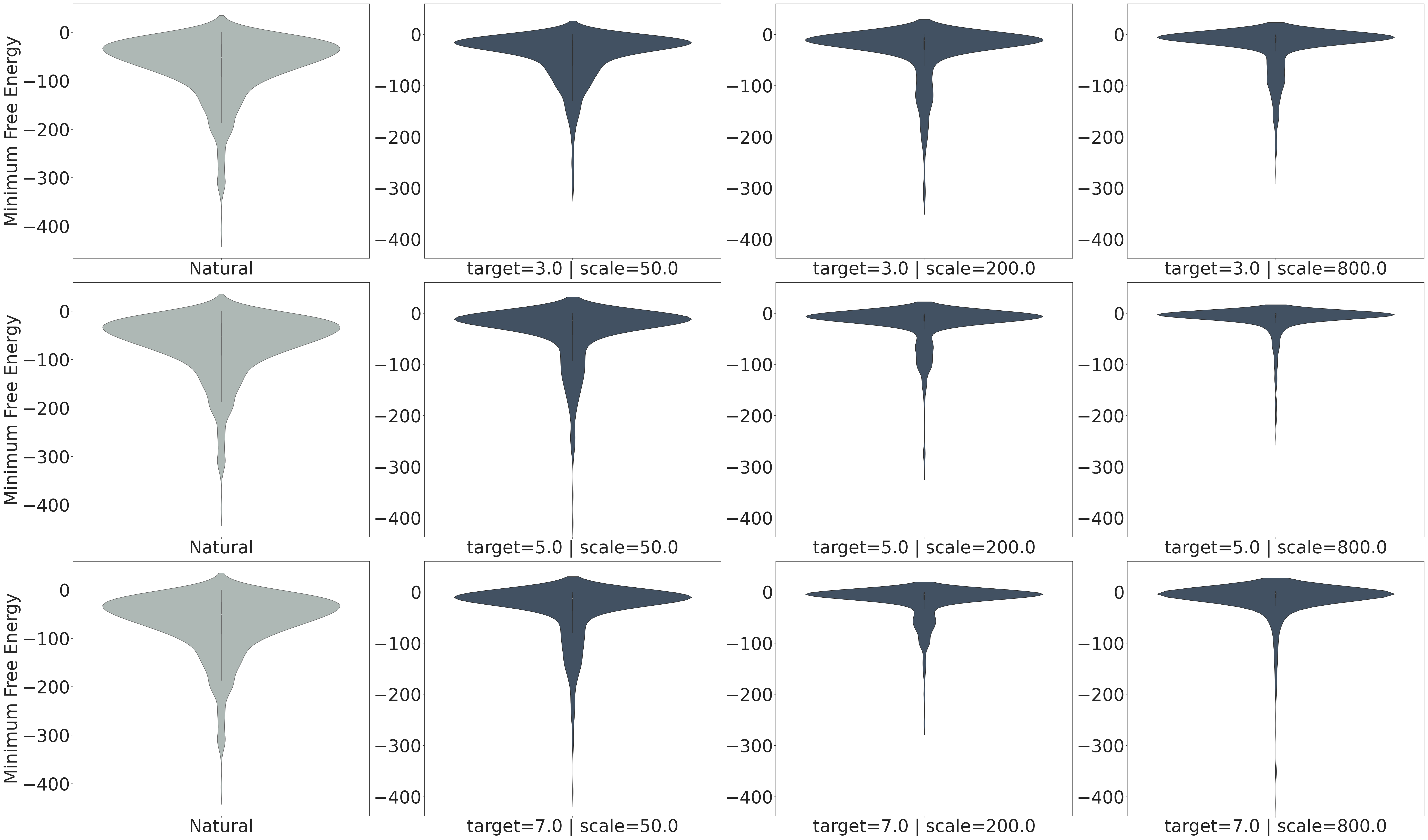}
}
\caption{Biological performance of the sequences generated with the MRL-guided diffusion model. The leftmost column shows the result of natural 5' UTR sequences. In the 2nd to 4th column, the guidance strength increases from 50 to 200 and 800. 
From top to bottom, each row corresponds to 
$\emph{log}_2$-MRL targets of 3, 5 and 7. 
} 
\label{fig:reward_optim_dist}
\end{figure}

\textbf{Biological Metrics.} We compare the generated distributions under different guidance strengths in terms of biological metrics (see Figure~\ref{fig:reward_optim_dist}). With the $\lambda$ and $r^*$ increasing, the disparities in G/C content and Minimum Free Energy (MFE) between the generated sequences and natural sequences become larger.
Specifically, the distribution of G/C content is shifting downwards while the mean MFE is increasing, indicating that the stability of generated sequences has been affected. 

\textbf{Reward Hacking in Reward-guided Diffusions.} As shown in Figure \ref{fig:reward_optim}, at a mild guidance strength, increasing the target TE or MRL values generally brings the reward improvements of generated sequences, demonstrating the efficacy of our reward-guided generation approach.
For higher guidance strength, the improvements can be \textit{artificially} amplified when evaluated with the same reward model used for guidance. Continuing to increase $\lambda$ and $r^*$ will eventually hurt the cross-validation performance, since the generation process is over-adapted towards the imperfect guidance reward model, which is a general phenomenon known as reward hacking~\citep{gao2023scaling,clark2023directly,zhang2024large}. 
Additionally, increasing $\lambda$ or $r^*$ may cause the distribution shift of biological properties of sequences such as G/C content and MFE (see Figure~\ref{fig:reward_optim_dist}), affecting the stability of UTR sequences. 
Thus, adopting moderate $\lambda$'s and $r^*$'s would be important for reward-guided generation in terms of the \textit{trade-off} between achieving higher reward values and maintaining generation qualities. 

\vspace{1em}
